**Building a Rich Dataset to Empower the Persian Question Answering Systems**

Mohsen Yazdinejad[1], Marjan Kaedi[2,*]

[1,2]Faculty of Computer Engineering, University of Isfahan, Azadi Sq., Hezarjarib St., Isfahan, Iran
*Corresponding author

**Abstract**

Question answering systems provide short, precise, and specific answers to questions. So far, many robust question answering systems have been developed for English, while some languages with fewer resources, like Persian, have few numbers of standard dataset. In this study, a comprehensive open-domain dataset is presented for Persian. This dataset is called NextQuAD and has 7,515 contexts, including 23,918 questions and answers. Then, a BERT-based question answering model has been applied to this dataset using two pre-trained language models, including ParsBERT and XLM-RoBERTa. The results of these two models have been ensembled using mean logits. Evaluation on the development set shows 0.95 Exact Match (EM) and 0.97 Fl_score. Also, to compare the NextQuAD with other Persian datasets, our trained model on the NextQuAD, is evaluated on two other datasets named PersianQA and ParSQuAD. Comparisons show that the proposed model increased EM by 0.39 and 0.14 respectively in PersianQA and ParSQuAD-manual, while a slight EM decline of 0.007 happened in ParSQuAD-automatic.

**1. Introduction**

The increase in the number of documents and text corpus, makes information retrieval (IR) a substantial issue [1]. IR systems extract users' desired information from documents using Natural Language Processing (NLP) [2]. One of the main tasks of NLP is Machine Reading Comprehension (MRC), which is capable of comprehending a text and helping Artificial Intelligence (AI) compete with humans. Developing such a system that understands textual data will be valuable because, in many tasks, people need to extract information from text [3].

Although, search engine is the application of IR techniques working based on keywords, users prefer to ask questions using natural language, and like to receive short and precise answers instead of long text. Therefore, researchers are trying to find another option for IR to take a step toward the smartification of current search engines, so that users can interact with computers by natural language. In this regard, Question Answering (QA) systems are considered an advanced form of IR [4] [5], which are a growing research field in the world [6], and demand for such systems is increasing day by day because they represent short, precise, and specific answers [7].

Search engines give some related and unrelated pages to users based on given keywords, while QA systems take natural language questions and retrieve proper answers automatically [8] [9] [10]

[11]. A reasonable answer is short, understandable, and accurate, which may be a word, sentence, paragraph, or complete document [12].

Generally, QA systems are classified into two major groups: open-domain and closed-domain. Suppose a QA system answers questions in a specific domain; for example, in health, it is called a closed domain. Closed-domain QA systems need knowledge in a particular field. Although a closed-domain QA system can give rapid, precise, and up-to-date answers in its specific domain, it can't be used in other areas [4].

Open-domain QA system is another kind of QA system used in different fields [4]. An open-domain QA system aims to extract answers from an extensive dataset. Existing datasets for reading comprehension mostly have one of the two following shortcomings:
- High-quality datasets [13] [14], are too small to train complex models.
- Large datasets [15] [16], are semi-artificial with low quality.

Therefore, to respond to the needs, a prominent and high-quality reading comprehension dataset, Stanford Question Answering datasets (SQuAD v1.0), has been proposed [17], which is freely available and consists of some Wikipedia articles. In this dataset, a part of the text is selected as the answer. SQuAD consists of 785,107 question-answer pairs in 536 articles and is almost two times bigger than previous reading comprehension datasets with manual labels like MCTest [13]. Some other large-scale annotated datasets are MCTest [18], Book-Test [19], SearchQA [20], NewsQA [21], ReCoRD [22], and ReCO [23], which have been recently represented.

So many robust QA systems have been specifically developed for English. Unfortunately, the primary QA datasets are in English, while languages with fewer resources need more work [3]. Considering Persian, a few pieces of research have been done, including PersianQA [24], ParSQuAD [25] and Rasayel&massayel [26]. ParSQuAD is a translation of SQuAD [17] in Persian using google-translate, and PersianQA is a small QA dataset including 10000 questions.

Despite the accessibility of different resources and datasets to answer retrieval, answering questions is still a challenging problem due to difficulty in comprehending questions and extracting answers [27]. As most existing systems support limited types of questions, to improve the performance of these systems, more efforts are required.

Since there is a limited number of datasets in Persian, this study has gathered a comprehensive and authoritative open-domain dataset for this language. Also, a Persian NLP module has been proposed to answer questions. The suggested method is an open-domain QA that can be used in all domains, such as medicine and sports.

The rest of this paper has been organized as follows: part 2, the related literature; part 3, the process of data collection and the methods used to make sure of a high-quality dataset; part 4, modeling; part 5, the evaluation of the gathered dataset, and part 6 conclusion and suggestions for future works.

## 2. Related Literature

Existing challenges in English and non-English QA models mostly rooted in quality and quantity of datasets. Therefore, studies about QA are divided into two parts: English and non-English datasets, and in this study, the focus in non-English is on Persian.

## 2.1. English Datasets

The increasing attention of researchers to MRC caused a growth in the significance of its datasets. Despite the fact that, there are many MRC datasets [18] [19] [17] [20] [21] [22] [23], there are still lots of features that can be considered to collect new datasets for MRC tasks.

MCTest [18] is one of the first and the smallest MRC datasets, including imaginary stories, and multiple choices questions. MCTest is too small for training big models.

SQuAD 1.1 [17] is another extensive dataset. It consists of a list of questions on a set of Wikipedia articles. The answer to each question is a part of the text (span), from the corresponding article.

SQuAD 2.0 [28] is generated using SQuAD 1.1, in addition to more than 50000 unanswered questions and some other answered questions. The presence of unanswered questions turns SQuAD 2.0 into a more challenging dataset. Existing studies on SQuAD 2.0 showed that MRC models performed better than humans.

SearchQA [20] is generated based on question-answer pairs from the popular television show Jeopardy. Each question has been googled, and retrieved text has been used as context. SearchQA consists of about 140000 question-answer pairs, and each has been associated with around 50 pieces of text.

ReCoRD [22] is automatically gathered from news articles. The discriminating characteristic of ReCoRD is that despite other MRC datasets like SQuAD, it needs common sense reasoning in several sentences. This article discusses the limitations of existing MRC models, which primarily rely on paraphrasing questions to find answers. To address this, the ReCoRD dataset was designed to minimize queries that can be answered merely by paraphrasing. Instead, ReCoRD focuses on requiring commonsense reasoning, distinguishing it from other datasets.

TWEETQA [29] is another detest in the question answering field, which has been produced from tweets that famous journalists have published. This dataset is suitable for question answering models that are looking to extract the answer for a question from an extensive text corpus.

## 2.2. Non-English Datasets

In the last few years, a high motivation is formed among researchers to collect low-resource languages datasets. Most of these datasets either use Wikipedia as their main resource, like SQuAD, or utilize the translation of the SQuAD, such as ReCo [23] in Japanese, ParSQuAD [25] in Persian, ARCD and Arabic-SQuAD [30] in Arabic, KorQuAD1.0 [31] in Korean, Spanish translation of SQuAD [32], FQuAD [33] in French, and a multilingual dataset [34]. In this study, the Persian MRC datasets are being elaborated more.

Razzagh Nouri et al. [35], represented a solution for question classification using machine learning approaches. There are three feature extraction methods illustrated in this study. In the first method, words are divided into clusters, and then feature vectors of questions are generated based on clustering information. The second method extracts features using a neural network. Each question turns into a vector, which is achieved using word2vec and is weighed by tf-idf coefficients. In the third approach, not only does it retain the novelty of the first approach, but also it takes into account data types that are sequence-based. In this paper, they introduced the UTQD.2016, collected from some Jeopardy games. The described feature vectors' efficacy is proofed by SVM and neural network results.

Veisi et al. [4] designed and implemented a system to answer medical questions in Persian. In this study, a collection of drug and disease documents have been gathered for a medical question answering. They converted these documents to a semi-structured form to extract answers easily. Their system has three modules: question processing, document retrieval, and answer mining. Sequential architecture has been designed for the question processing module, which retrieves the central concept of questions using different approaches: rule-based methods, NLP, and dictionary-based methods. In the document retrieval module, the text corpus is listed and searched through the Lucene Library. The retrieved documents have been identified using similarity algorithms, and the highest rank of documents is chosen for use in the answer extraction module. The most related part of the retrieved document is found using this module. The results demonstrate that this system performances excellently in answering various questions about diseases and drugs.

In another study, Ben Veyseh [36], introduced a new interlingual approach employing integrated semantic space among languages. After keyword extraction, entity linking, and detecting the type of answer, lingual reciprocal semantic similarity is used to find answers from the knowledge base through selecting relation and adaptation. This approach has been evaluated for Persian and Spanish. The results were hopeful for both languages and leading to introducing a new QA dataset for Persian. He added the Persian language to the QALD-5 dataset, which is a multilingual dataset. The questions were translated into Persian by a language expert from outside the development team, to add Persian translation to these datasets. They used majority voting among five annotators, to annotate keywords of each Persian question.

In another study, Boreshban et al. [26] discuss the development of a Persian QA corpus, named Rasayel and Massayel dataset. The corpus includes a number of non-factoid and factoid questions, each annotated with details like question type, difficulty, and expected answer type. It's a valuable resource for learning various aspects of question answering systems, including question classification and answer extraction.

Mohtashami and Sarmadi [37], have detected the type of question raised in Persian sentences with the help of convolutional neural networks and LSTMs. The methods used have achieved acceptable accuracy. Considering the requirement of a dataset which includes sentences and their labels for training classification models, the UTQD.2016 was opted. Since in this kind of dataset, the type of question is not labeled, the labeling of this dataset has been manually developed in eight classes.

Mohammadi et al. [38] proposed a method based on categorizing and weighting words for answering biographies in Persian, and made a small dataset. Technical terms in this field were determined by using a corpus of answers, which were manually extracted. Then, these terms were weighted and used for categorizing the questions and the candidate sentences as answers. The questions in this study have been limited to questions about "date of birth", "date of death", "place of birth", "place of death", and or a combination of them.

Farahani et al. proposed a BERT for Persian (ParsBERT). In this study, an extensive dataset for some NLP tasks was collected, and a pre-trained model was created. ParsBERT has outperformed multilingual BERT and other earlier works in Text Classification, Sentiment Analysis, and NER across all datasets [39].

XLM-RoBERTa [40] is a relatively new and big interlingual language model based on RoBERTa and has been trained on 100 languages on CommonCrawl filtered by 2.5 TB. Unlike other XLM models, XLM-RoBERTa doesn't need language code to comprehend language and can determine the correct language from input identifiers. This is a powerful tool for understanding multilingual language and useful for solving ambiguities in a multilingual setting.

PersianQA [24] is a public dataset produced by Sajjad Ayoubi. This dataset is among the first QA datasets in Persian, which has smoothened the path for researchers in this type of study. This dataset is in the format of SQuAD 2.0 and contains more than 9,000 records in train set and 900 records in development set.

PCoQA introduces a dataset specifically designed for Persian conversational question answering, containing 9,026 questions in dialogical form. The dataset focuses on enhancing performance through historical queries and serves as a foundational resource for Persian conversational QA systems [41].

PQuAD is a Persian question-answering dataset aimed at improving NLP capabilities for the Persian language. It features 80,000 questions sourced from Persian Wikipedia and is utilized as a benchmark for model evaluation, addressing unique challenges in Persian QA [42].

MeDiaPQA provides a specialized dataset for medical question answering in Persian, comprising 15,748 question-answer pairs across over 70 medical specialties. It facilitates domain-specific modeling and QA applications in healthcare [43].

IslamicPCQA focuses on complex, multi-hop reasoning questions derived from Islamic textual resources. It includes 12,282 question-answer pairs and is inspired by frameworks like HotpotQA to enhance Persian QA capabilities in complex reasoning tasks [44].

Since there is not much perfect and standard dataset in Persian question answering, and there is room for improving the existing dataset, an ideal dataset for Persian is presented in this study by gathering standard and open-domain data to enhance the performance of question answering in Persian.

## 3. The Proposed Dataset Gathering

Having a robust QA system, a complete and standard dataset is required. Considering shortcomings in Persian datasets, a complete and comprehensive dataset called NextQuAD is introduced in this study. Figure 1 shows the proposed approach for gathering NextQuAD. Data is collected with three methods:
- Translation of SQuAD 2.0
- "Quiz of Kings" question-answer pairs
- Manual data gathering from some well-known Persian websites by searching on Google.

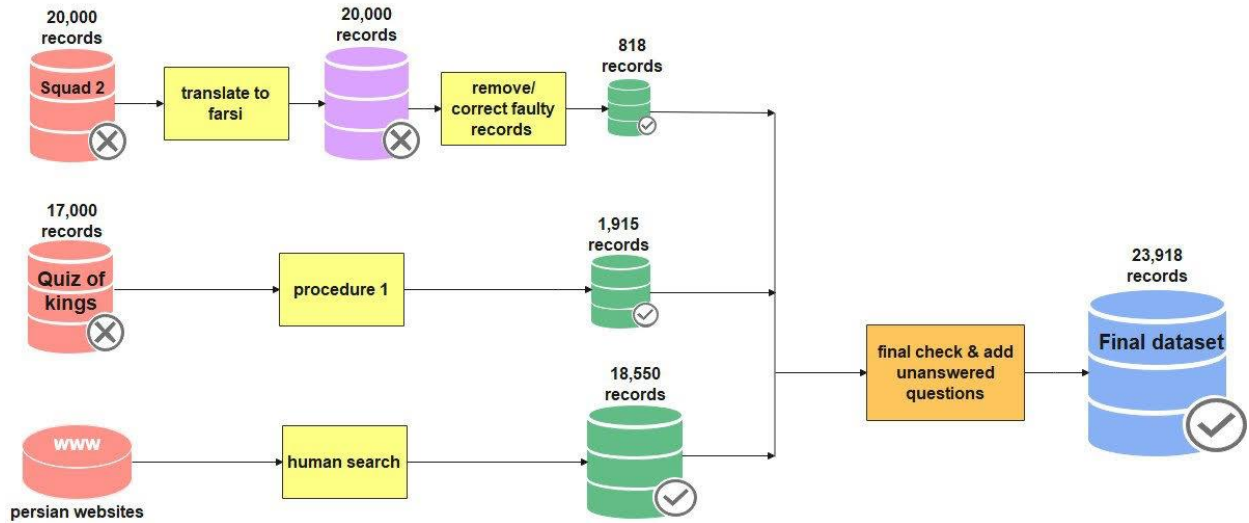

Figure 1) Proposed dataset gathering

### 3.1. SQuAD 2.0 Translation

In the first step of gathering the NextQuAD, SQuAD 2.0 is translated into Persian. To translate the dataset, Google Translation API [45] is used in python which leads in having more than 20 thousand interpreted records. From this, after undergoing a manual checking process, only 845 items have suitable translations or are useful with a bit of alternation. As this process is expensive in terms of human force and time, and the quality of most of the translated records is low, the process is stopped and we leave translation of the remained records of SQuAD 2.0.

### 3.2. "Quiz of Kings" question-answer pairs

In this step, like in SearchQA [18], a Persian game application called "Quiz of Kings", in which some questions are asked of users, and their knowledge is challenged, was used. This part of our dataset is produced using 17,000 questions and their answers on this application, by applying a web scraping technique through the following process:
- Concatenating questions and answers, and searching them in Google.
- Choosing the first founded result and opening the related webpage.
- Calculating the similarity between paragraphs of this webpage and the searched text, using BERT [46].
- Selecting the most similar paragraph as the context of that question. If this paragraph has a similarity under 0.8, this question-answer pair is ignored. This threshold has been determined through a process of trial and error.
- Finding the answer in the context and using its index for start and end.

Questions and answers have existed before, and the above process has been only used to find context and span. All this process resulted in 1,915 records.

### 3.3. Manual Data Gathering

Using the previous two steps, less than 3,000 accurate and high-quality records were collected. Next, seven crowdworkers helped to produce new question-answer pairs in different fields to complete the dataset, which are named in Table 1. Exploration was conducted through one of the following valid websites: Persian Wikipedia[1], Chetor[2], Namnak[3], Digiato[4], Zoomit[5], and some other news websites. This process leaded to production of more than 21,000 records in 4 months.

### 3.4. Final Dataset Preparation

After merging the aforementioned datasets, 7,515 contexts and 23,918 pairs of question and answer were achieved. Table 1 shows the frequency of contexts for each subject.

Table 1) Frequency of each subject in NextQuAD

| Subject | Frequency in Train Set | Frequency in Development Set | Total Frequency |
|---|---|---|---|
| Householding | 1,301 | 139 | 1,440 |
| Politics | 1,019 | 107 | 1,126 |
| Art | 759 | 70 | 829 |
| Sport | 623 | 71 | 694 |
| Economy | 585 | 57 | 642 |
| Psychology | 562 | 67 | 629 |
| Science | 541 | 52 | 593 |
| Health | 414 | 53 | 467 |
| Literature | 345 | 32 | 377 |
| History | 266 | 28 | 294 |
| Tourism | 173 | 27 | 200 |
| Religion | 127 | 12 | 139 |
| Geography | 73 | 12 | 85 |

As depicted in Table 2, 10% of data was kept for the development set.

Table 2) Frequency in train and development set

|  | Train set data | Development set data |
|---|---|---|
| Number of contexts | 6,788 | 727 |
| Number of answered questions | 14,847 | 1,586 |
| Number of unanswered questions | 6,757 | 728 |
| Total number of questions | 21,604 | 2,314 |

---

[1] https://fa.wikipedia.org/wiki/%D8%B5%D9%81%D8%AD%D9%87%D9%94_%D8%A7%D8%B5%D9%84%DB%8C
[2] https://www.chetor.com/
[3] https://namnak.com/c1-%D8%A7%D8%AE%D8%A8%D8%A7%D8%B1
[4] https://digiato.com/
[5] https://www.zoomit.ir/

As in this dataset there are some cases with the same context assigning to multiple question and answer pairs, to prevent data leakage, we ensure to have each context with this feature entirely whether in train set or in development set.

For each question, several answers were considered in the development set (like SQuAD 2.0). For example, if a text is about Ali Daei, and the question is "How many national goals does Ali Daei have?", the answers could be "109", "109 goals", or "109 national goals". Therefore, it is necessary to have two or three answers for some questions in the development set. Figure 2 presents their frequency.

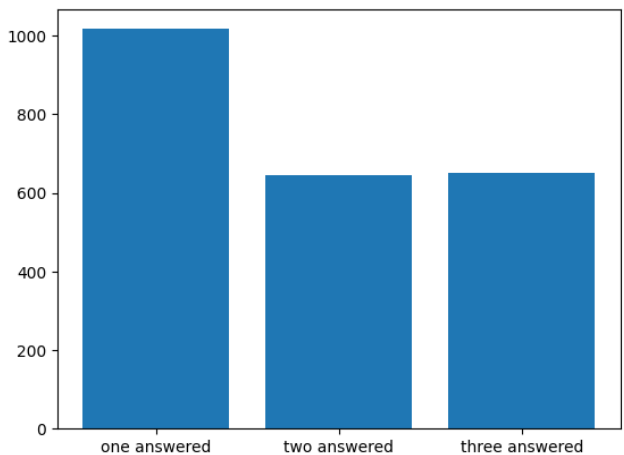

**Figure 2) Frequency of questions with one/two/three answers in the development set**

### 3.5. Inter Annotator Agreement

To determine the Inter-Annotator Agreement (IAA), we evaluate a subset of the dataset with three metrics: Exact Match (EM), Jaccard score, and F1 score.

Firstly, we select 50 labels from each crowdworker randomly, in the next step, allocate them to remained crowdworkers for collecting 7-way annotation. So, they have been provided with 300 pairs of content and question. Finally, while they repeatedly perform the annotation process, they do not have any information about their peers' answers.

The aforementioned steps conclude in having 350 pairs, in which each question will have seven answers with various ways of annotation which can be seen in Table 3.

**Table 3) Inter Annotator Agreement sample and its translation**

|  | Truncated Content | Question | Answer 1 | Answer 2 | Answer 3 | Answer 4 | Answer 5 | Answer 6 | Answer 7 |
|---|---|---|---|---|---|---|---|---|---|
| **Persian** | ماقبل تاریخ یا پیشا تاریخ به دوره قبل از تاریخ مکتوب اطلاق می شود . در پی مهاجرتهای اولیه بشر در عصر... | انسان های مدرن چند سال پیش برای اولین بار به اروپا رسیدند؟ | 40,000 سال | 40,000 سال | 40,000 سال پیش | 40,000 | 40,000 سال | 40,000 سال | 40,000 سال |

| English Translated | Prehistory refers to the period before written history. Following the early human migrations in the era … | How many years ago did modern humans first arrive in Europe? | 40,000 years | 40,000 years | 40,000 years ago | 40,000 | 40,000 years | 40,000 years | 40,000 years |
|---|---|---|---|---|---|---|---|---|---|

For calculating IAA, the proposed methodology by Karpukhin et al. [47] was utilized, in which the answer of the main annotator considered as ground truth and the rest of answers assumed as predictions. Next, the maximum for each metrics was calculated over our six predictions, and average each of them over 350 questions. The results have been shown in table 4. As all these results are more than 0.8, this guarantees having a reasonable annotation [48].

**Table 4) Results of three metrics in IAA**

| EM | Jaccard | F1 |
|---|---|---|
| 0.961 | 0.977 | 0.974 |

### 3.6. Dataset Texts Normalization

While there are some characters in Persian and some in Arabic which can be employed interchangeably, they have different unicodes. It makes words with such letters to be treated as two different words. So, standard character forms must be used in datasets [49]. In this step, all Arabic characters in the dataset are substituted with their Persian forms.

### 4. Modeling

The NextQuAD was evaluated using two pre-trained transformer-based language models, ParsBERT and XLM-RoBERTa. As we had limitations in GPU memory, the highest length for each context was considered 310 words, and K-fold cross-validation with five folds was employed to evaluate the modeling results on the training dataset.

There are some basic models (trained in each fold) that outputs of their softmax output layer are two vectors with the length of 310, which estimate the place of start and end. Thus, according to [50] and [51], the results of basic models were ensembled to improve their performance. Mean logits was used as an ensemble method. The result of the ensemble method for the start point is the average of start vectors achieved by basic models, and precisely similar to that, the result for the end point is the average of end vectors performed by basic models.

### 5. Experiments

As there are five folds, and in each of them, an independent model is trained, at the end of each fold, its model is saved and evaluated on the development set[6]. As it can be perceived in table 5, XLM-RoBERTa has outperformed ParsBERT.

---
[6] NextQuAD dataset and its related Python codes are pushed in https://github.com/mosiomohsen/NextQuAD

Table 5) Results of model evaluation on five folds

| Model | ParsBERT | | | XLM-RoBERTa | | |
|---|---|---|---|---|---|---|
| Metric | Jaccard | EM | F1_score | Jaccard | EM | F1_score |
| Fold 1 | 0.8636 | 0.8064 | 0.8740 | 0.9722 | 0.9494 | 0.9717 |
| Fold 2 | 0.8631 | 0.8103 | 0.8727 | 0.9693 | 0.9460 | 0.9693 |
| Fold 3 | 0.8562 | 0.8008 | 0.8663 | 0.9662 | 0.9425 | 0.9665 |
| Fold 4 | 0.8597 | 0.8042 | 0.8701 | 0.9643 | 0.9365 | 0.9652 |
| Fold 5 | 0.8687 | 0.8086 | 0.8795 | 0.9720 | 0.9494 | 0.9716 |

The ensemble method has been implemented in three approaches:
- Ensemble of 5 models, which are the results of different folds of XLM-RoBERTa.
- Ensemble of 5 models, which are the results of different folds of ParsBERT.
- Ensemble of all ten above models.

Table 6 shows the EM, F1_score, and Jaccard of three ensemble approaches.

Table 6) Results of three ensemble approaches

| | Jaccard | EM | F1_score |
|---|---|---|---|
| Ensemble of 5 XLM-RoBERTa models | **0.9754** | **0.9542** | **0.9748** |
| Ensemble of 5 ParsBERT models | 0.8684 | 0.8124 | 0.8780 |
| Ensemble of all ten models | 0.9434 | 0.9225 | 0.9415 |

Considering table 6, the ensemble of XLM-RoBERTa models has the best performance.

Table 7 shows the model's results on answered and unanswered questions separately.

Table 7) Results of the model to answered and unanswered questions separately

| | Jaccard | EM | F1_score |
|---|---|---|---|
| Answered questions | 0.9717 | 0.9407 | 0.9708 |
| Unanswered questions | 0.9835 | | |

To compare the quality of NextQuAD with other Persian datasets, the trained model with this dataset was applied to the development set of ParSQuAD and PersianQA datasets. Table 8 shows the results of our ensemble model as well as the performance of their models.

Table 8) Comparison of our proposed model on other datasets with their main models

| Dataset | ParSQuAD-automatic | | ParSQuAD-manual | | PersianQA | |
|---|---|---|---|---|---|---|
| Metric | EM | F1_score | EM | F1_score | EM | F1_score |
| Their model | 0.6773 | 0.7084 | 0.5286 | 0.5666 | 0.5210 | 0.7860 |
| Our model | 0.6705 | 0.6951 | 0.6698 | 0.6945 | 0.9065 | 0.9557 |

According to table 8, the performance of the model trained by NextQuAD is better than the other two models. Since the same algorithm and language model have been used in these three studies, only the difference in dataset quality could be the reason for this superiority in results. Some influencing factors in making such a comprehensive and complete dataset are: the strict selection of SQuAD 2.0 machine translation and manually correcting its mistakes, using different subjects in producing the dataset, giving help from different people with varying points of view and tastes, dataset cross-checks between crowdworkers and correcting personal mistakes, and a final review and correction by one person.

**6. Conclusion**

QA system is a growing research field, and demand for this kind of system has been increasing. QA systems have been well developed in different languages, especially English. Nonetheless, as there are rare rich datasets for Persian, there would be a blockage in developing a robust QA system for this language. So, this study gathered and designed a comprehensive and authoritative open-domain dataset for the Persian called NextQuAD. This dataset was collected through three approaches: 1- translation of SQuAD 2.0, 2- "quiz of kings" question-answer pairs, and 3- manually gathered data from valid Persian websites. 23,918 questions and answers were gathered in 7,515 contexts, 10% of which were kept for evaluation in the development set, and the rest were used in the training set.

The model trained by NextQuAD is applied to the development parts of ParSQuAD and PersianQA to compare the quality of NextQuAD and the other two datasets. The results showed that our proposed model performs better than the two other models.

Comparing NextQuAD with ParSQuAD shows that a large dataset doesn't necessarily improve the quality of the model, and the dataset quality is more important. Developing and completing NextQuAD, both in the case of subjects and questions diversity, can improve QA models in Persian. Also, considering the particular form of QA models' outputs, it seems that there are lots of innovative ensemble techniques that can boost the results.

**Availability of supporting data**

All materials of this research are available in github:
https://github.com/mosiomohsen/NextQuAD
The dataset will be added after our research is published. If that's required for reviewers, we could send it.